\documentclass[letterpaper, 10 pt, conference]{ieeeconf}  

\IEEEoverridecommandlockouts                              

\usepackage{microtype}
\usepackage{graphicx}
\usepackage{dblfloatfix}
\usepackage{cite}
\usepackage{booktabs} 
\usepackage{amsmath, amssymb, bm}
\usepackage{algorithmicx, algpseudocode, algorithm}
\usepackage{multicol, lipsum, caption}
\usepackage{comment}
\usepackage{subcaption}

\usepackage{multirow}

\title{\LARGE \bf
A Hierarchical Approach to Active Pose Estimation
}

\author{Jascha Hellwig$^{*}$, Mark Baierl$^{*}$, Jo\~{a}o Carvalho$^{\dagger}$, Julen Urain$^{\dagger}$, Jan Peters
\thanks{$^{*\dagger}$Equal contribution.}
\thanks{All authors are with the Intelligent Autonomous Systems, Technische Universit\"{a}t Darmstadt, Germany - \{jascha.hellwig, mark.baierl\}@stud.tu-darmstadt.de, \{joao,julen\}@robot-learning.de, jan.peters@tu-darmstadt.de}
\thanks{Correspondence to joao@robot-learning.de}
}

\usepackage{xcolor}
\usepackage{todonotes}

\newcommand{\jascha}[1]{\todo[inline,color=green!30]{Jascha: #1}}

\newcommand{\statespace}{\mathcal{S}}
\newcommand{\actionspace}{\mathcal{A}}
\newcommand{\observationmodel}{\mathcal{O}}
\newcommand{\statetransitionmodel}{\mathcal{T}}
\newcommand{\observationspace}{\Omega}
\newcommand{\rewardfunction}{\mathcal{R}}
\newcommand{\policy}{\pi}
\newcommand{\policytree}{\pi_{\text{tree}}}
\newcommand{\policyrollout}{\pi_{\text{rollout}}}

\newcommand{\nrsimulations}{N_{\text{sim}}}

\newcommand{\nrparticles}{K}

\newcommand{\state}{s}
\newcommand{\action}{a}
\newcommand{\observation}{o}
\newcommand{\img}{\mathbf{I}}
\newcommand{\pointcloud}{\pmb{PC}}

\newcommand{\history}{h}
\newcommand{\reward}{r}

\newcommand{\envMap}{\mathbf{M}}
\newcommand{\posAgent}{\mathbf{p}^{agent}}
\newcommand{\posObject}{\mathbf{P}^{object}}
\newcommand{\observationVector}{\mathbf{\observation}}

\newcommand{\blackboxsim}{\mathcal{G}}

\newcommand{\currentaction}{\action_{t}}

\newcommand{\currentobservation}{\observation_{t}}
\newcommand{\currenthistory}{\history_{t}}
\newcommand{\currentbelief}{\mathcal{B}(\history_t)}
\newcommand{\currentparticles}{B^i_t}

\newcommand{\nextaction}{\action_{t+1}}

\newcommand{\Reals}{\mathbb{R}}

\newcommand{\expectation}{\mathbb{E}}

\begin{document}

\newpage
\newpage

\maketitle
\thispagestyle{empty}
\pagestyle{empty}

\begin{abstract}
Creating mobile robots which are able to find and manipulate objects in large environments is an active topic of research.
These robots not only need to be capable of searching for specific objects but also to estimate their poses often relying on environment observations, which is even more difficult in the presence of occlusions.
Therefore, to tackle this problem we propose a simple hierarchical approach to estimate the pose of a desired object.
An Active Visual Search module operating with RGB images first obtains a rough estimation of the object $2$D pose, followed by a more computationally expensive Active Pose Estimation module using point cloud data.
We empirically show that processing image features to obtain a richer observation speeds up the search and pose estimation computations, in comparison to a binary decision that indicates whether the object is or not in the current image.





\end{abstract}


\section{Introduction}

A major task of autonomous robots working in manipulation scenarios is to find objects and estimate their poses.
While in recent years there have been major developments in pose estimation methods~\cite{li2021leveraging,Kleeberger2021SingleShot,Zhang_2021_CVPR,Lin2021DualPoseNet}, these assume the object was already found and is present in the viewing frustum.
However, the object might be occluded or even out of the field of view. 
In these situations, the agent might be required to navigate along and interact with the environment to improve its estimation of the object's pose.
We call Active Pose Estimation (APE) to the problem of actively deciding how to interact with the world in order to minimize the uncertainty with respect to the pose of a target object. 
In most cases, to estimate the pose of an object, we are required to sense depth information by using depth cameras or Lidars.
Nevertheless, the computational requirements of dealing with depth information are high.
We remark that in APE, we are expected to decide the next action every step, and thus, we aim to have low computational requirements to increase the control frequency.



We call Active Visual Search (AVS) to the problem of actively deciding how to interact with the environment to search and find a particular object.
A possible solution to increase the control frequency is to run a search process with RGB information only and then, once the object is found, run a pose estimation method that uses point clouds, such as the Iterative Closest Point Algorithm (ICP) \cite{icp}.
While this can lead to good performance in terms of pose estimation, it might not be very robust.
If the search procedure terminates without a decent view of the object, the pose estimation quality could decrease significantly. 
To improve robustness, it feels a more natural solution to apply an active pose estimation process at the expense of higher computation.



With the aim of having a both computationally light and at the same time robust pose estimator; in this paper, we propose a novel algorithm for occluded or out-of-view object pose estimation.
We propose a hierarchical approach that combines a computationally light AVS process with a computationally intense APE process. 
To avoid the high computational budget of estimating an object pose in high-dimensional point clouds, we combine an AVS running on RGB observations with an APE using pointcloud inputs.
The search module uses RGB images to efficiently interact with the environment in order to estimate a rough two-dimensional object pose. 
When the belief in the object's 2D pose, computed with the search module, is sufficient, we pass that belief to the pose estimation module and actively move to estimate the 6D pose of the object.
We frame both process in terms of a POMDP problem and solve it by adapting the Partially Observable Monte-Carlo Planning (POMCP) algorithm~\cite{pomcp} to the problem of pose estimation.
Framing both process as POMDP permits sharing the computed belief distributions in the AVS process to the APE process, reducing the computational complexity for the second. 


In particular, the main contributions of these work are
(i) We propose a unified framework that combines a 2D AVS process with a 3D APE process. Due to the unified framework, we present an approach to transfer the beliefs states from the AVS to the APE process. (ii) We propose two approaches to apply POMCP for both 2D active search and 3D active pose estimation. (iii) We make an extensively ablation study to investigate the performance of our proposed method under multiple hyperparameters. We investigate the required control steps until the pose is properly estimated.

\begin{figure*}[!ht]
    \centering
    \includegraphics[width=0.9\textwidth]{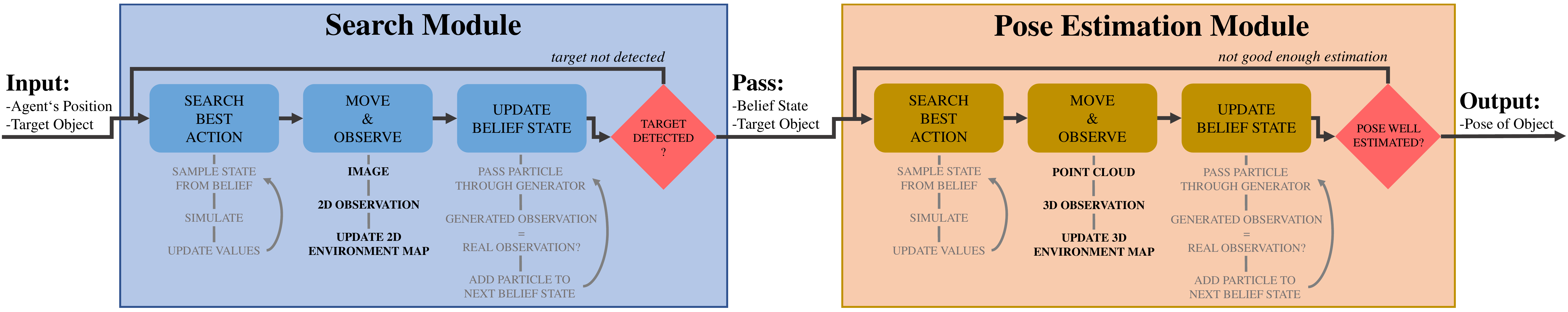}
    \caption{POMCP Search \& Pose Estimation Modules. 
    In the Search Module a target object is searched within a 2-dimensional space using RGB images, producing a belief state of the object's pose in $2$D.
    This belief acts a prior for the Pose Estimation Module, where the full pose is computed using point clouds.
    }
    \label{Modules_Combined}
    \vspace{-0.4cm}
\end{figure*}


\section{Related Work}

AVS is commonly framed as solving a Partially Observable Markov Decision Process (POMDP).
In general POMDP-based approaches use observations to update the environment map and compute optimal actions using the current map estimate.
POMCP-based Online Motion Planning (POMP)~\cite{pomcpAVS} knows the environment map before hand and uses the POMCP algorithm~\cite{pomcp} to select optimal actions.
Unobserved map cells are marked as \textit{candidate} and observed cells marked according to the observation. 
The agent explores candidate zones until the object is found, which is determined by a binary object detector that can have errors.
A robust visual docking mechanism compensates object detection errors, by returning the agent to a known position if the object said to be found was not encountered.
POMP++~\cite{pomcp++AVS} extends POMP to the case where the map is unknown and is dynamically created, which enlarges the search space to navigate the environment.
\cite{continuous_pomcp_find_people} presents two approaches that extend the general POMCP algorithm to the continuous domain, Continuous Real-time POMCP and Adaptative Highest Belief Continuous Real-time POMCP Follower, which are validated in real-life scenarios by navigating over $3$km.
\cite{semantic_obj_search} uses semantic environment models in combination with various search strategies, and does not use the POMCP algorithm.


In contrast to POMDPs, deep RL methods learn a parametrized policy to navigate through environment. This allows faster control frequencies in deployment, as there is no any planning phase.
Generalizable Approaching Policy LEarning (GAPLE)~\cite{gaple} uses depth information for policy learning, whereas~\cite{SemExp} builds an episodic semantic map of the environment to explore efficiently.
\cite{action_obj_perceiver} combines two deep neural network modules, one that learns to detect given objects in its view and another that learns to approach the target object. 
\cite{crfAVS} uses a Conditional Random Field (CRF) to build a prediction map representing the most promising locations for the searched object.
While these approaches showed good performance, they require a large amount of training data to be able to generalize to unseen situations, and one of the reasons for focusing on solving POMDPs instead.





\begin{figure*}[!ht]
    \centering
    \subfloat[Promising Believed Object Position]{%
      \includegraphics[width=0.3\textwidth]{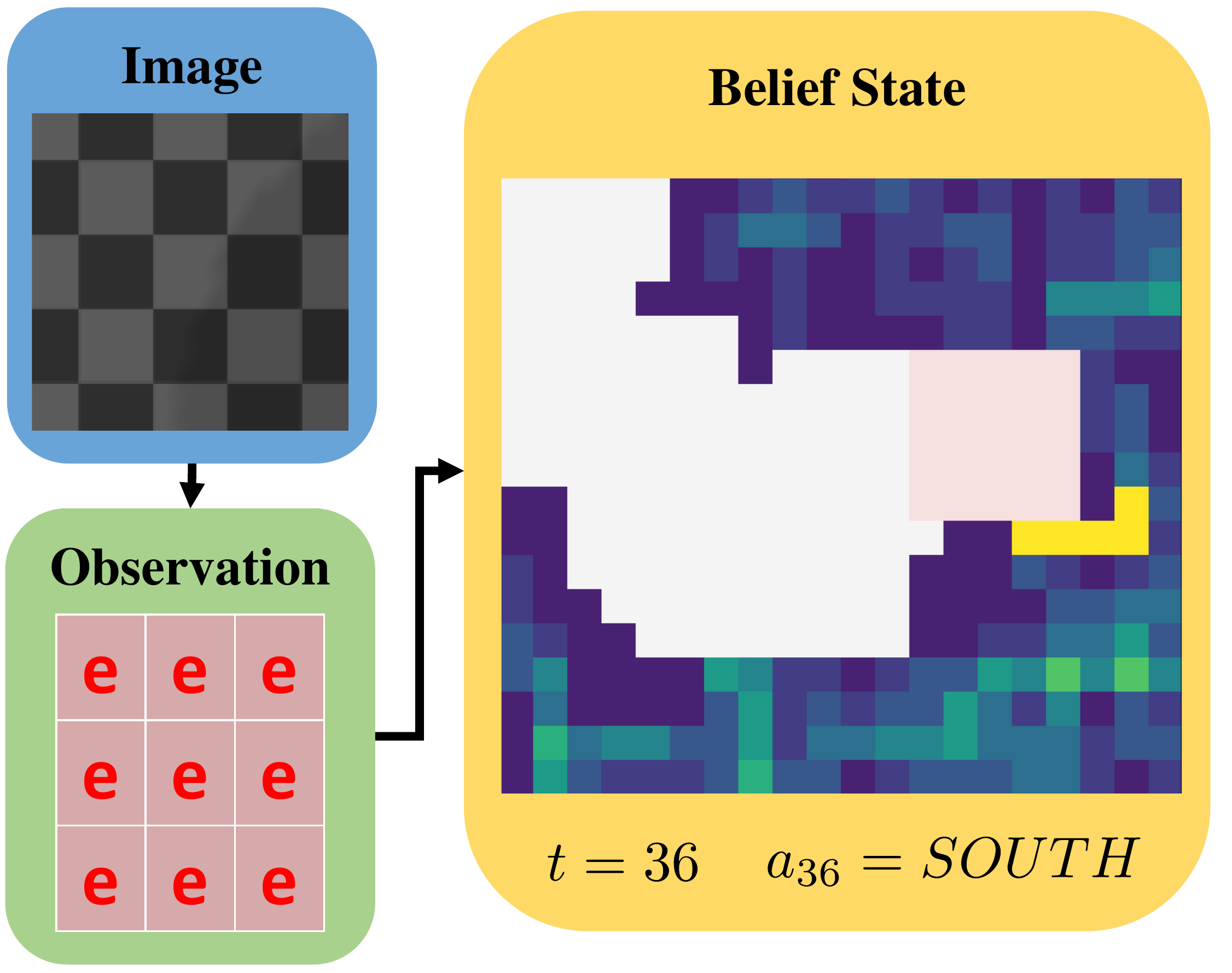}%
      \label{belief_state_updates_1}%
    }\qquad%
    \subfloat[Object Not Found - Resample Belief State]{%
      \includegraphics[width=0.3\textwidth]{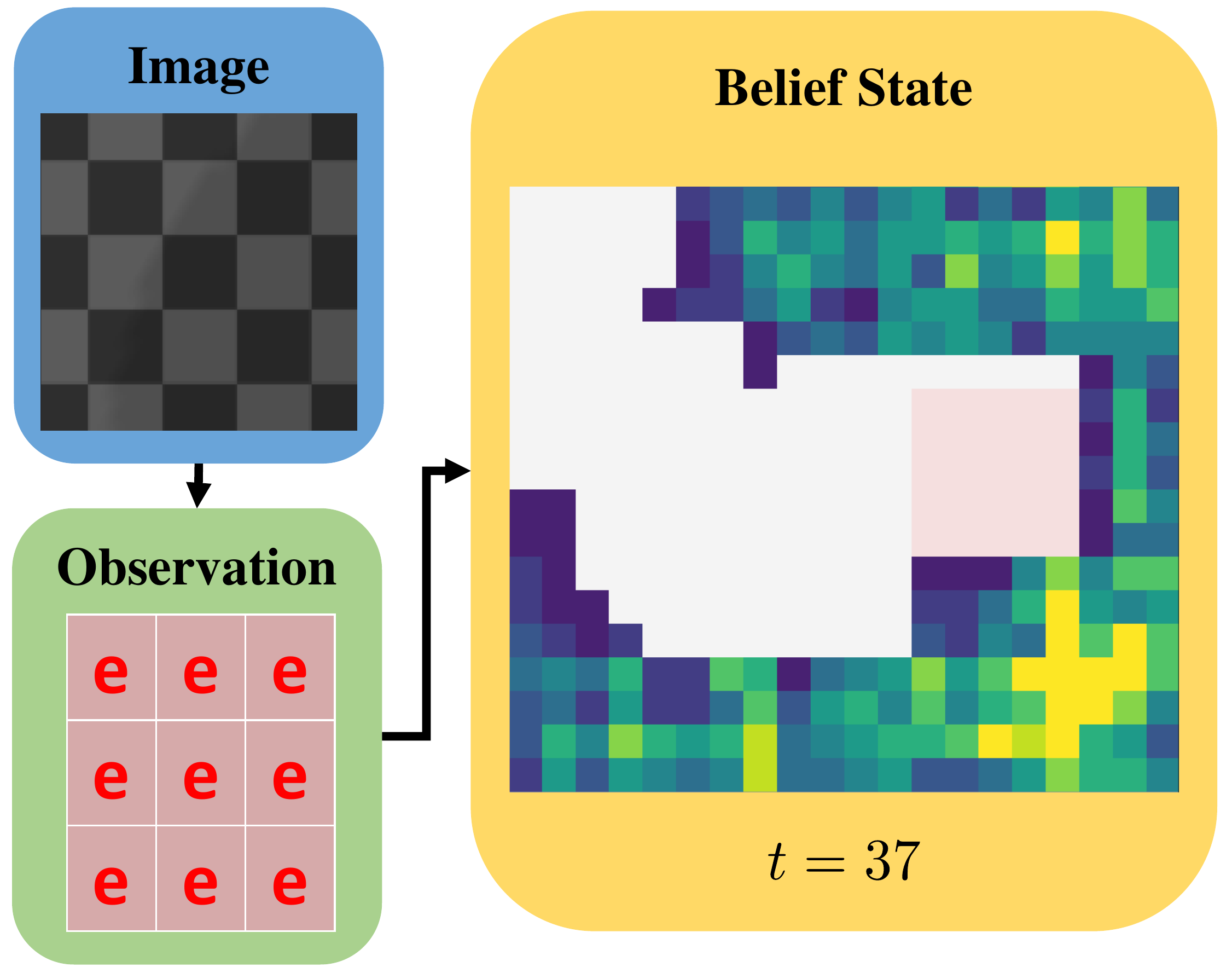}%
      \label{belief_state_updates_2}%
    }\qquad%
    \subfloat[Object Found - Particles Agree]{%
      \includegraphics[width=0.3\textwidth]{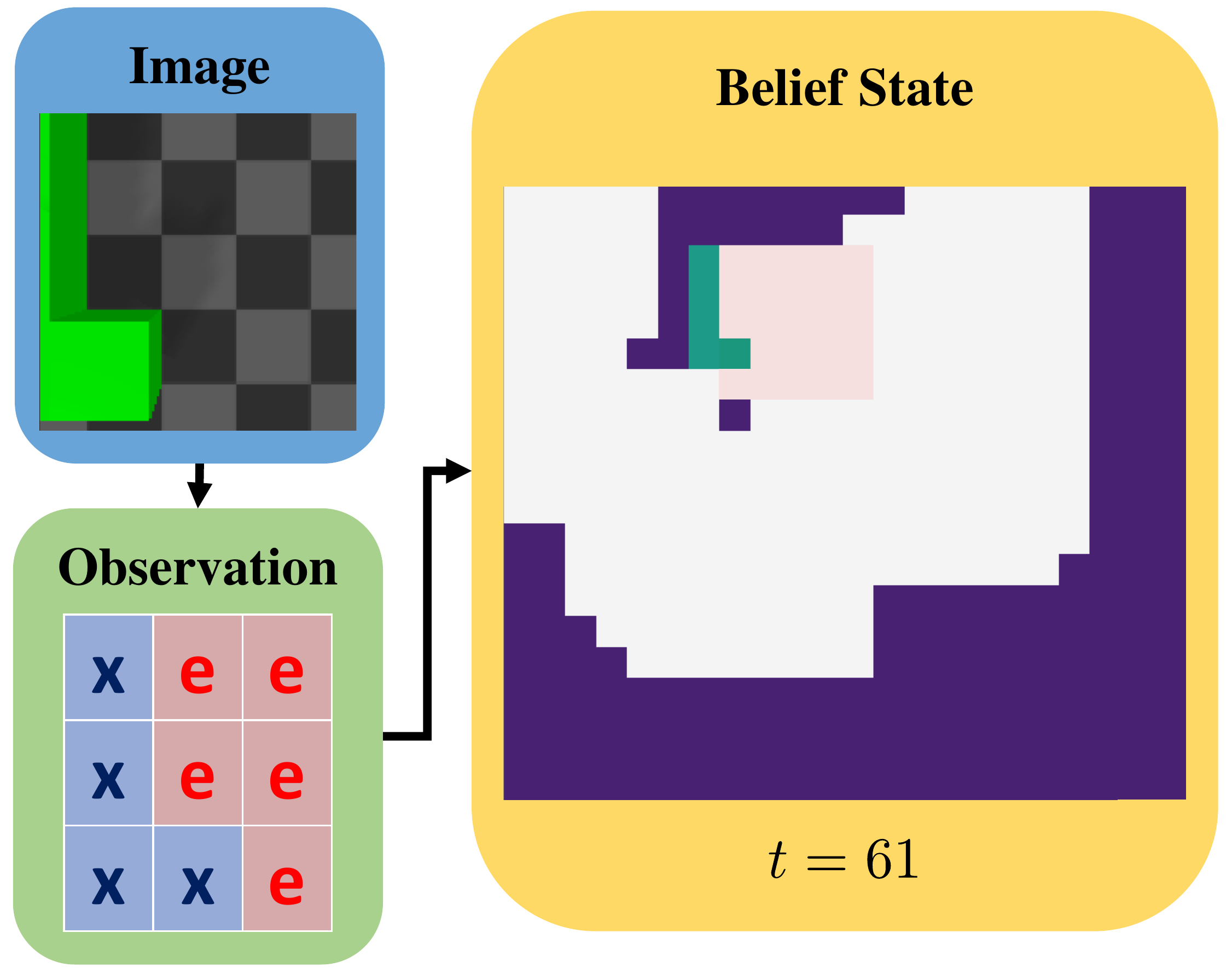}%
      \label{belief_state_updates_3}%
    }
    \caption{
    Illustration of the belief and object position over time steps (indicated with $t$).
    The light pink square represents the current agent's position.
    Parts of the map the agent has already observed are marked in white.
    a) At time-step $t=36$ many particles agree on the object $2$D pose, given by the marked yellow \textit{L} shape, and therefore the agent moves $SOUTH$.
    b) Because the object is not the location the agent believed at $t=36$, particles are resampled, resulting in a new belief state.
    c) At time-step $t=61$ the agent finds the object and all particles agree.
    }
    \label{belief_state_updates}
\end{figure*}

\section{Background}

Our proposed method is framed as a POMDP problem. Then, we solve the POMDP problem by adapting the POMCP algorithm to the pose estimation problem. In the following, we introduce the POMDP problem and we briefly explain the POMCP. 

\subsection{Partially Observable Markov Decision Processes}
A POMDP~\cite{pomdp} is described as a tuple $\langle \statespace, \actionspace, \statetransitionmodel, \rewardfunction, \observationspace, \observationmodel, \gamma \rangle$, where $\statespace$ is a finite set of states, $\actionspace$ a finite set of actions, $\statetransitionmodel: \statespace \times \actionspace \to \statespace$ is a state transition model, $\rewardfunction: \statespace \times \actionspace \to \Reals$ is a reward function, $\observationspace$ is a finite set of observations, $\observationmodel: \statespace \times \actionspace \to \observationspace$ is an observation model, and $\gamma \in [0, 1)$ is a discount factor.
At time step $t$ an agent is in state $\state_t$, performs an action $\action_t$ according to its policy $\policy$, obtains an immediate reward $\reward_t$, transitions to state $\state_{t+1}$ and obtains an observation $\observation_{t}$.
All the variables are available to the agent, expect for the internal state $\state$, which it cannot observe.
Therefore, and opposite to a Markov Decision Process (MDP), the policy is conditioned on the history of previous interactions with the environment and not on the current state $\policy(\action_t \mid \history_{t})$, where $\history_t=\{\observation_0, \action_0, \ldots, \observation_{t-1}, \action_{t-1} \}$.
Since the true belief state for a history $\currenthistory$ is not known, it is approximated using particles $\currentparticles$. Each particle corresponds to a sample state and the sum of all particles builds the current belief state $\currentbelief$.
The goal of a planning agent is to obtain an optimal policy to maximize the expected sum of discounted rewards ${J_{\policy} = \expectation_{\policy, \statetransitionmodel, \observationmodel} \left[ \sum_{t=0}^{\infty} \gamma^t \rewardfunction(\state_t, \action_t) \right]}$.

\begin{table}[t]
    \centering
    \begin{tabular}{lc}
         \textbf{Symbol} & \textbf{Meaning}  \\
         \hline \\
         $s_t$ & state at time $t$ \\
         $a_t$ & action at time $t$ \\
         $\mathbf{o}_{s_t,a_t}$ & observation for state $s_t$ and action $a_t$\\
         $\envMap$ & environment map \\
         $V$ & observed environment state \\
         $\mathbf{p}^{agent}$ & agent's state \\
         $\mathbf{P}^{object}$ & object's state belief \\
         $\mathbf{I}$  & observed RGB image \\
         $\mathbf{PC}$ & observed pointcloud \\
         $\mathcal{G}_{\mathbf{I}}$  & simulation environment \\
    \end{tabular}
    \caption{Used Notation.}
    \label{tab:my_label}
    \vspace{-0.4cm}
\end{table}

\subsection{Partially Observable Monte-Carlo Planning}



A widely used online planning method for POMDPs is the POMCP algorithm~\cite{pomcp}. It extends Monte-Carlo tree search (MCTS)\cite{mcts} to POMDPs and combines it with a Monte-Carlo update of the agent's belief state.
POMCP addresses the intractability of online planning in large POMDPs by utilizing Monte-Carlo sampling for belief state updates and planning. In addition it requires only a black-box simulator of the POMDP, rather than explicit probability distributions, in order to sample state transitions and observations.
This algorithm can be divided in two major steps:
\begin{itemize}
    \item[I)] Select the best action using an extension of the UCT algorihtm~\cite{UCT} for POMDPs - Partially Observable UCT.
    \item[II)] Update the belief state after an action is performed and an observation is received, using Monte-Carlo belief state updates.
\end{itemize}

\noindent\textit{I) Partially Observable UCT}:
To select the action, three procedures are used: 
\begin{enumerate}
    \item \textsc{Search} - search for the best action.
    \item \textsc{Simulate} - determine the value of a state.
    \item \textsc{Rollout} - extend the search tree.
\end{enumerate}

The \textsc{Search} procedure starts with sampling a state $\state$ from the current belief $\currentbelief$.
$\state$ and the history $\currenthistory$ is given to the \textsc{Simulate} procedure in order to evaluate the value of different actions.
This process is repeated $\nrsimulations$ times before selecting the next action to update the search tree and approximate the  action values.

The \textsc{Simulate} procedure uses two different policies to traverse the search tree, depending on whether the given history is already present in the search tree or not. 
If the history is present, a tree policy $\policytree$ is used to select actions while navigating through the search tree. 
This tree policy uses an extension of the UCB1 algorithm~\cite{ucb1} for POMDPs in order to compute the value of performing an action given the current history. 
The best action is then passed to the black box simulator $\blackboxsim$ to generate an observation, a reward and the next state $(s',o,r) \sim \blackboxsim(\state, \action)$.
The next state $s'$ is used to recursively call the \textsc{Simulate} procedure and accumulate the returns in a discounted fashion until a terminal state or a maximum depth is reached. 
If the history is not present during the \textsc{Search} procedure, a new node $T(ha)$ in the tree is created for each possible action. 
Then the \textsc{Rollout} procedure is called to determine the value of the given history. 

The \textsc{Rollout} procedure selects an action according to a random rollout policy $\policyrollout$ that is given to the black box simulator to generate an observation, a reward and the next state. 
The next state is then used to recursively call the \textsc{Rollout} procedure, until a terminal state or a maximum search depth is reached and a discounted reward is accumulated.

Finally the real action to be executed by the agent $\currentaction$ can be selected using the values returned by the \textsc{Search} procedure and a real observation $\currentobservation$ is received, which is used with Monte-Carlo belief state updates to update the belief state before selecting the next action $\nextaction$.

\noindent\textit{II) Monte-Carlo Belief State Update}:
In order to update the belief state, $\nrparticles$ particles are used. 
Each particle $\currentparticles$ represents a randomly selected state $s$ from the current belief state $B_t$,
and is then passed with the selected real action $\currentaction$ to the black box simulator to receive an observation $\observation$ and the next state $\state'$. 
The observation $\observation$ is used to check, whether it matches the real observation from the environment $\currentobservation$. 
If they match $\observation = \currentobservation$, the state $s'$ is added to the next belief state. 
This process is repeated until $\nrparticles$ particles are added to the next belief state $B_{t+1}$.


\begin{figure*}[!ht]
    \centering
    \includegraphics[width=\textwidth]{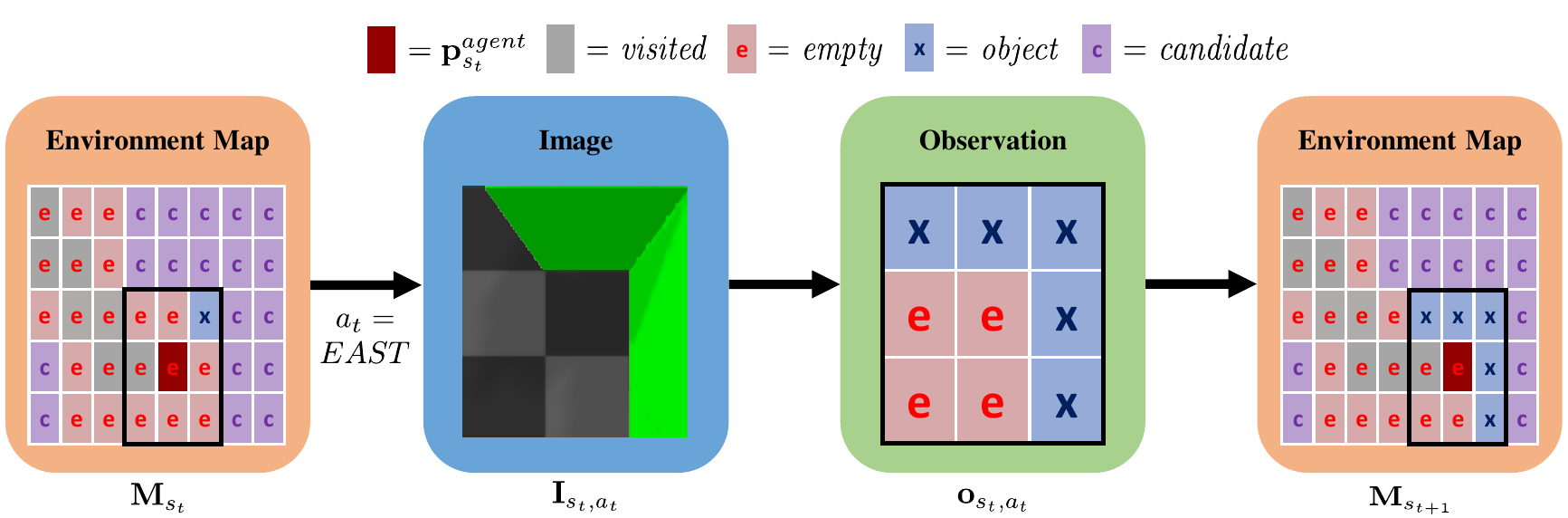}
    \caption{
    Environment map update between two steps $\envMap_t \rightarrow \envMap_{t+1}$ in the 2D Search Module. 
    The leftmost figure shows the current map and position of the agent (center of black borders).
    After executing the real action $a_t$ in the current state $s_t$, an RGB image $\img_{s_t,a_t}$ is received (2nd figure), which is processed into an observation (3rd figure).
    With this observation the agent updates his internal map of the environment accordingly.
    Notice the area inside the black borders.
    }
    \label{Image_To_Obs_To_Env}
\end{figure*}

\section{Combining Active Visual Search\\and Active Pose Estimation in POMDPs}
The problem we address with our method is the one of estimating the pose of an object in an environment using visual information. This means that the agent has no access to the true state of the environment, but to observations. 

We consider the situation in which the object might not be in our field of view at first, but it might be occluded or even in a different area of the environment.
To deal in a computationally efficient way with this problem, we split the process in two stages. First, 
the agent navigates along the environment in order to get a decent view for estimating its pose. We call to this stage, the Search stage and we apply AVS methods to search efficiently. Second, once the object is found, we apply APE methods to estimate the pose of the object. Namely, we call it Pose Estimation stage.
An overview of our method that combines those two stages, is shown in Fig.~\ref{Modules_Combined}.

AVS has been shown to work with RGB images and therefore we do not use point cloud data when searching the object. Because of this, we obtain a computationally more efficient approach.
Hence, we merge AVS using RGB images and APE using point clouds into a single approach.
While it is possible to first use AVS and then do pose estimation independently, in our approach we use the environment information from the search module as a prior for the pose estimation module improving the computational efficiency. In the following, we introduce extensively the algorithm we run on each module.

\subsection{Search Module} \label{search_module}
We solve the AVS problem with POMDPs methods. 
We propose to solve the search problem in a two-dimensional projection of the environment. This projection has been previously shown to be enough~\cite{pomcpAVS}. 
We stick to this approach and therefore define our environment map as $\envMap \in V^{x \times y}$ where $V$ are the possible values of a cell and $x,y$ the discrete cell position in the environment. $V$ can take $5$ possible values, representing the knowledge we have about the observed environment, ${V:=\{empty, candidate, object, other\_object, blocked\}}$.
Finally, for each time step $t$ the agent is in a specific state $s_t$, which represents both the knowledge of the environment and the belief states of our search object. 
This state is a tuple $s_t=(\envMap, \posAgent, \posObject)$, where $\envMap_{s_t}$ is the state's map of the environment, the agents position $\posAgent_{s_t} \in \mathbb{N}^2$  and the believed object position $\posObject_{s_t} \in \mathbb{N}^{obj\_size \times 2}$. 
Our definition of the believed object position allows to define objects corresponding to the occupied cells in the environment map. 
Therefore, each object is described by locations in $\envMap$ according to the level of discretization. 
For example, a cube might take a single cell whereas an edge-like shape might occupy three cells.

\begin{figure*}[!ht]
    \centering
    \includegraphics[width=\textwidth]{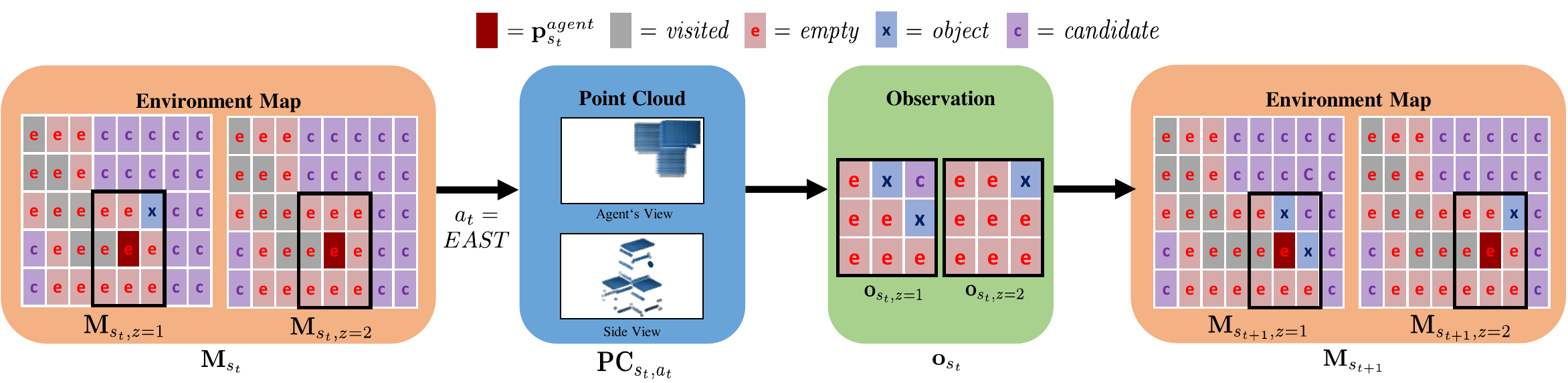}
    \caption{
    Environment map update between two steps $\envMap_t \rightarrow \envMap_{t+1}$ in the 3D Pose Estimation Module.
    The leftmost figure shows two levels of the current map ($z=1$ and $z=2$) and position of the agent (center of black borders).
    After executing the real action $a_t$ in the current state $s_t$, a point cloud $\pointcloud_{s_t,a_t}$ is received (2nd figure), which is processed into an observation $\observationVector_t$.
    The observation has the $Z$-levels, where each is used to update the corresponding level in the environment $\envMap_{t,z=i} \rightarrow \envMap_{t+1,z=i}, i=1,2,\text{...},Z$.
    With this observation the agent updates his internal map of the environment accordingly.
    Notice the area inside the black borders.
    }
    \label{PointCloud_To_Obs_To_Env}
\end{figure*}

\begin{figure}[!h]
    \begin{center}
    \centerline{\includegraphics[width=1.0\columnwidth]{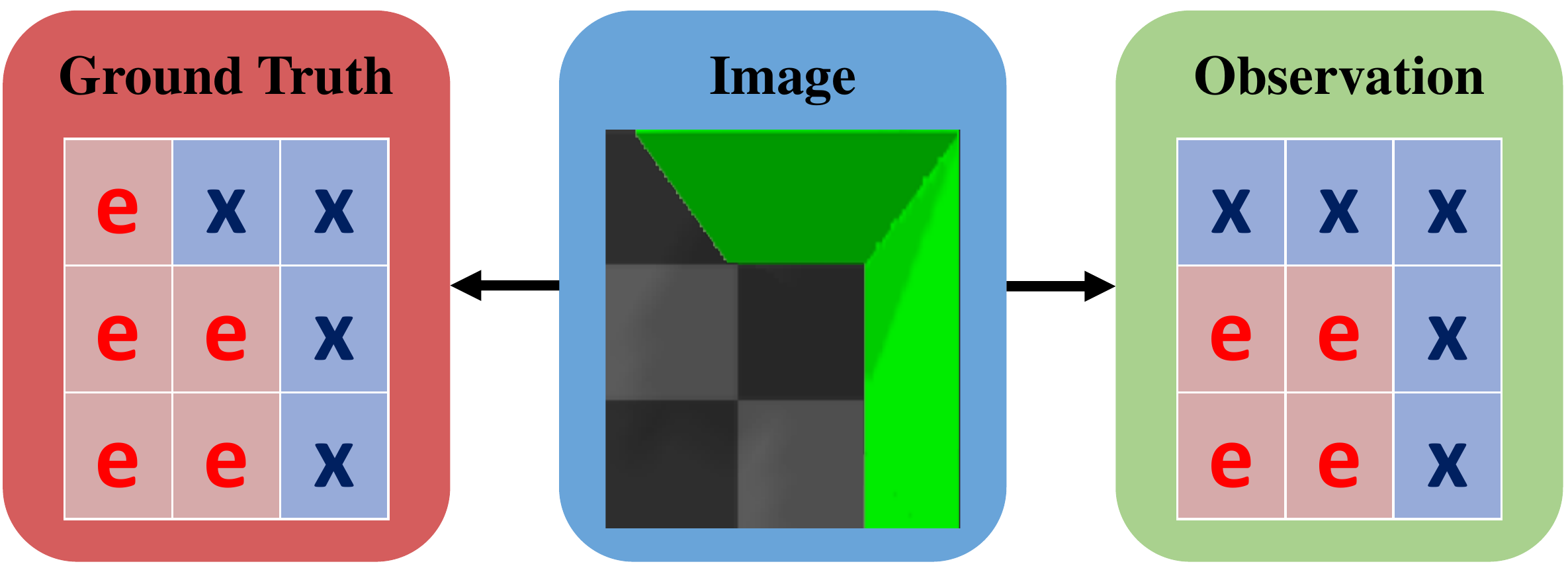}}
    \caption{
    Mapping from RGB images to observations.
    Creating observations from images can lead to mistakes, e.g. due to perspective distortion.
    The upper left corner in the observation is marked with \textit{object}, while the ground truth is \textit{empty}.
    }
    \label{groundtruth_image_observation}
    \end{center}
\end{figure}

In a POMDP, observations are the only way an agent can reason about the state of the environment. 
In our case we use RGB images to produce observations which are then used to update the gathered information about the environment, as depicted in Fig.~\ref{Image_To_Obs_To_Env}.
We use observations to replace $candidate$ values of the environment map with observed values. 
Those observed values are generated by transforming an image $\img_{\state_t,\action_t}$ to an observation $\observationVector_t$ after an action $\action_t$ in state $\state_t$ is performed. 
The observation $\observationVector_t \in V^{w \times h}$ represents, in which cells of the image the searched object is observed, where $w$ is the amount of  cells observed along the x-axis and $h$ the y-axis ones.
Various methods can be used to transform an image into this type of observation, e.g. color based decisions can be sufficient. 
Those observations are introducing noise and misidentified cells as can be seen in Fig.~\ref{groundtruth_image_observation}. 
However, we can show that they are expressive enough and allow to successfully find objects.

Most important, those type of observations allow the agent to reason about the environment and the object by himself. 
This means if an object is identified in some part of the image, the agent will explore that region until it observes the full object. 
This exploration is the major benefit towards the binary observations mentioned earlier.
There is more information that is passed to the agent and therefore used for action selection. These observations allow to move to believed object positions in the received images, rather than simply exploring until the object is found in an image. 

Since our method is using POMCP, a black box simulator $\blackboxsim_{\img} (\state_t, \action_t)$ is required to perform simulations. 
It returns for a given state $\state_t$ and action $\action_t$ an successor state $\state'$ and the corresponding generated observation $\observationVector_{s_t, a_t}$ along with the reward $\reward_t$. 
Since, we cannot generate images $\img_{\state_t,\action_t}$ in order to generate the observation $\observationVector_{s_t, a_t}$, the black box simulator has to be capable of surpassing the need for images. 
We solve this by directly creating the observation $\observationVector_{s_t, a_t}$ by generating it from the environment map $\envMap_{s_t}$ of the given state $\state_t$. 
We know the position and orientation of the agent as well as the characteristics of the camera used to take images, which allows us to calculate in which parts of the image, an object should be seen. 
It is important to notice, that this calculation is independent of the true position of the searched object, but only dependent on the believed object position $\posObject_{s_t}$ of the state $\state_t$. 
This means, after a true observation from the environment is received, we can pass all particles to the black box simulator to generate observations. 
Those generated observations are used to discard the particles which believed in a contradicting object position. 
For example, if a particle beliefs the object to be seen, when taking the action $a_t$, but taking that action in the environment results not in observing the object, the particle is thrown away and a new particle is sampled.

Using those generated observations to update the belief state results in more correct believed object positions, which will finally result in finding the object at the believed object position, as can be seen in Fig.~\ref{belief_state_updates}. 
A terminal state $s_t$ has $\forall (x,y) \in \posObject_{s_t} : \envMap_{s_t}(x,y) = object$. 
This can be interpreted as a state, whose believed object position matches its past observations from the environment. 
When such a terminal state is reached, the current belief state in terms of particles is passed to the pose estimation module and the search module terminates.

\subsection{Pose Estimation Module}
The pose estimation module follows the same concept as the search module, with two major differences: (1) the environment map is now three-dimensional: $\envMap \in V^{x \times y \times z}$, and (2) the observations are produced using point clouds $\pointcloud_{s_t,a_t}$. 

In order to use the prior belief state, each particle $B^i_{2D}$ from the search module corresponds to a particle $B^i_{3D}$ in the Pose Estimation module. 
The particles in the Pose Estimation module now set the lowest level of their environment map to the one passed from the search module ${\forall x \forall y : \envMap_{B^i_{3D}}(x,y,0)=\envMap_{B^i_{2D}}(x,y)}$ and all other levels are filled with $candidate$ values. 
Therefore, the initial belief represents a rough belief of the object's pose and the three-dimensional environment in general. 

In order to estimate a more precise pose, we use point clouds $\pointcloud_{s_t,a_t}$ instead of images $\img_{s_t,a_t}$ and change the observation to be three-dimensional $\observationVector_t \in V^{w \times h \times d}$, where $d$ is now representing the observed cells of the environment along the $z$-axis. 
The update process for the environment map $\envMap_{s_t}$ after an observation $\observationVector_t$ is received, is the same as in the case of the search module, but now with respect to three-dimensions, as can be seen in Fig.~\ref{PointCloud_To_Obs_To_Env}.

Using point clouds introduces a different type of noise as images do. 
\begin{figure}[!h]
    \begin{center}
    \centerline{\includegraphics[width=1.0\columnwidth]{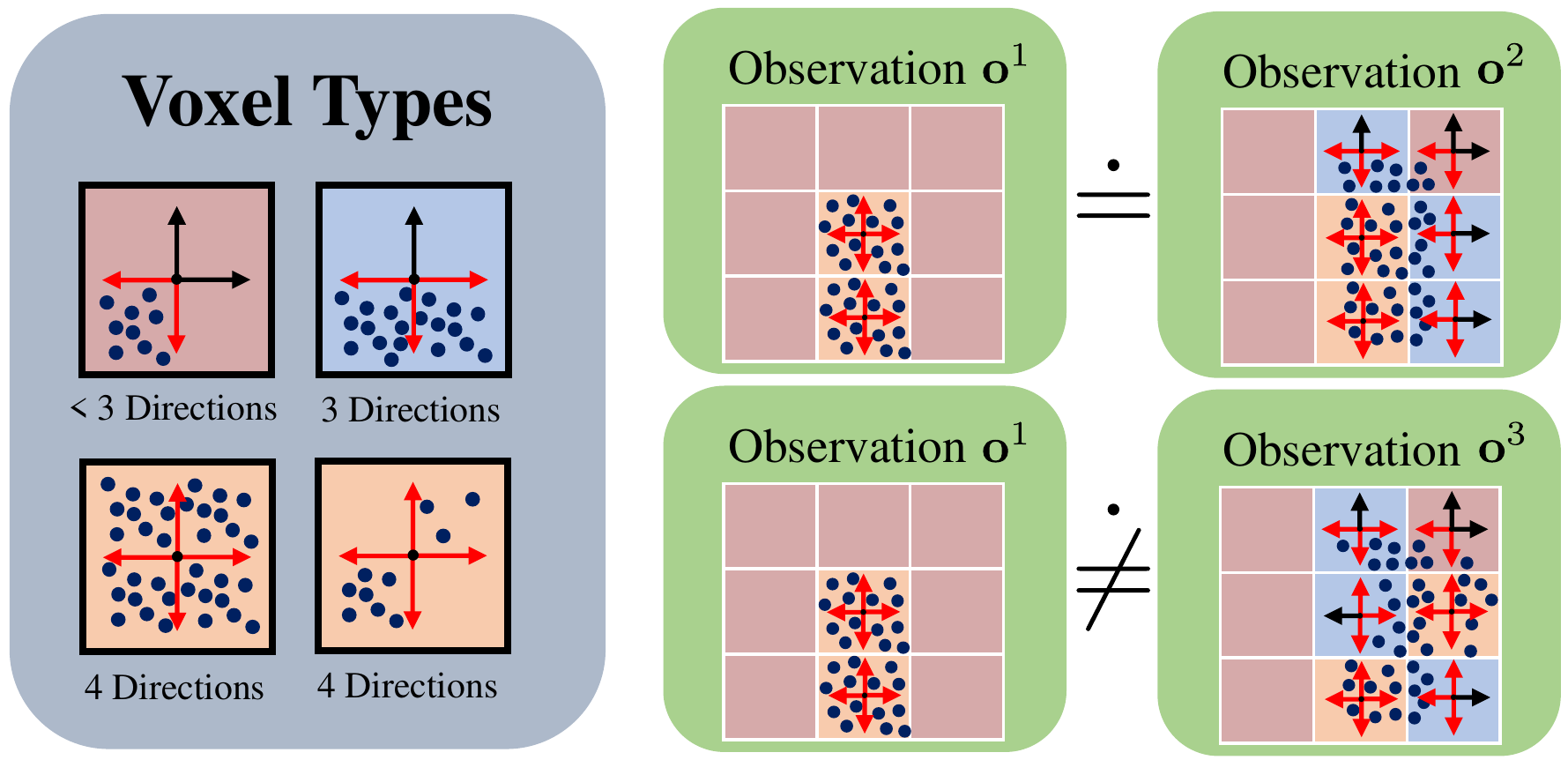}}
    \caption{Soft equality for point cloud observations.
    Cells with points in all quadrants (orange) are denoted \textit{core} cells.
    }
    \label{soft_equality}
    \end{center}
\end{figure}
In order to handle noisy observations we introduce a soft equality ($\dot=$) relationship. 
We classify every cell in a level of the environment map $\envMap_{s_t}$ as one specific type according to the positioning of the points it contains, before assigning an observed value to it.
In more detail, only cells classified as \textit{core} cells (see Fig.~\ref{soft_equality}), are used for testing Soft Equality of two observations. This means only observations with equal core cells are considered as equal.

\begin{figure}[ht]
    \begin{center}
    \centerline{\frame{\includegraphics[width=0.65\columnwidth]{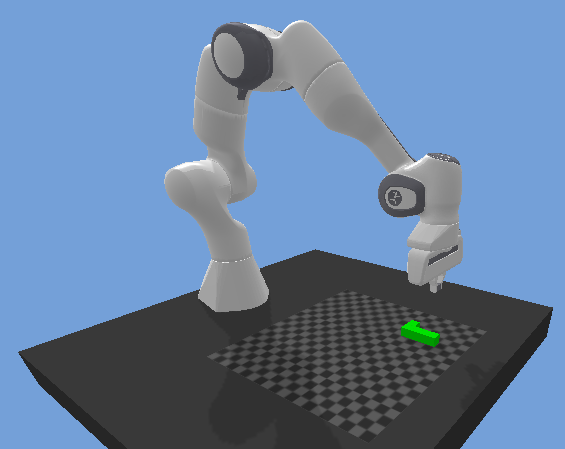}}}
    \end{center}
    \caption{
    Simulation environment with the Franka Emika Panda robot arm in PyBullet.
    A virtual camera is attached to the end-effector to capture RGB images.
    }
    \label{env_img}    
\end{figure}
\section{Experiments}

A new black box simulator $\mathcal{G}_{\pointcloud}(s_t,a_t)$ is introduced, which follows the same intuition of the simulator $\mathcal{G}_{\img}$, but now generates point clouds according to the belief in the object's position and so generates three-dimensional observations. 
This procedure can take more time than when using an RBG image, but also gives a more precise pose estimation of the object.



Our experiments were designed to answer the following questions:

\begin{enumerate}
    \item Does our mapping from images to observations improve over a binary observation decision in the context of AVS?
    \item Can the combination of search with a pose estimation module estimate object poses in a POMDP?
    \item How is our approach affected by hyperparameters and different types of objects?
\end{enumerate}

\begin{figure}[ht]
    \begin{center}
    \centerline{\frame{\includegraphics[width=0.76\columnwidth]{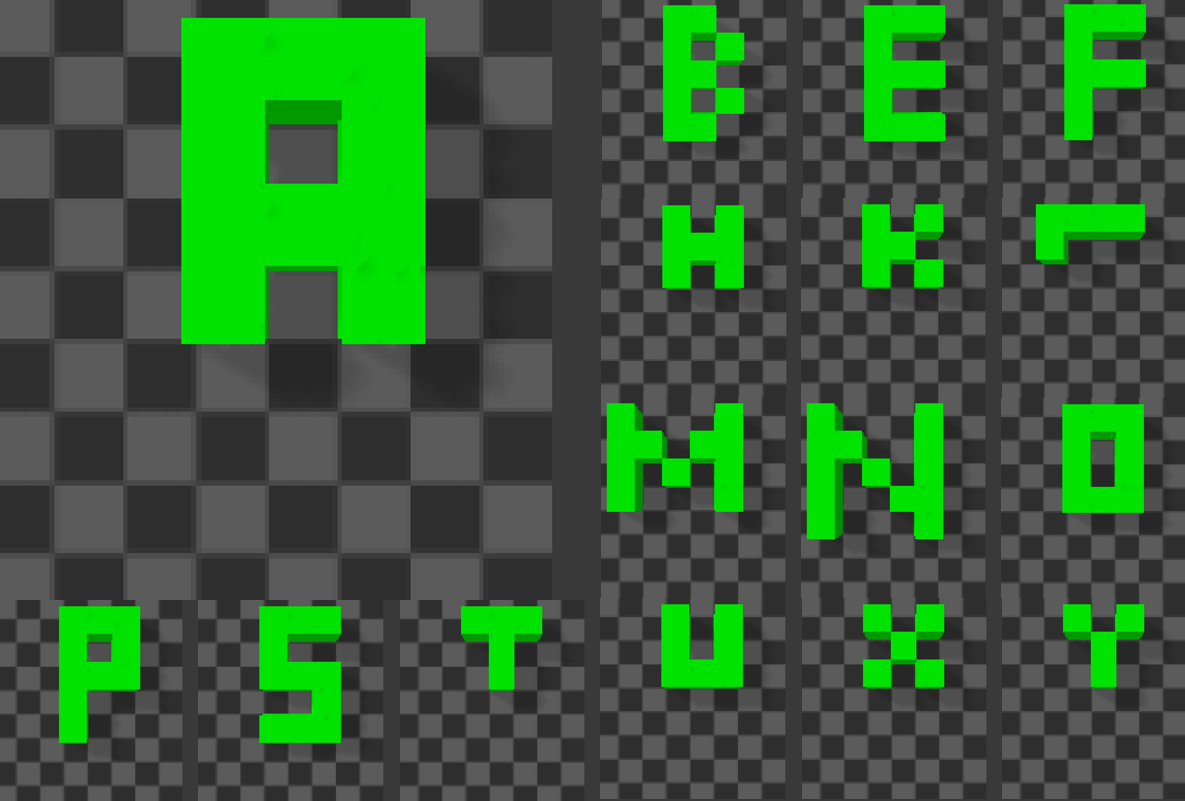}}}
    \end{center}
    \caption{
    Examples of objects considered in the experiments.
    }
    \label{letters_img}    
\end{figure}

Our environment contains an agent depicted as a Franka Emika Panda robot arm that has to search for objects laying on a table as in Fig.~\ref{env_img}.
The implementation uses PandaGym~\cite{pandagym} and Pybullet~\cite{pybullet}. 
The search space is defined by a $20 \times 20$ grid on the table with $5$ levels in the 
$z$-dimension. 
The considered objects are letters represented by multiple connected cubes placed in the grid, as depicted in Figure \ref{letters_img}.
The action space is $\actionspace = \{NORTH, EAST, SOUTH, WEST\}$, which corresponds to moving the end-effector in discrete steps along the $x$- and $y$-directions while maintaining a fixed height $z=5$. 
Observations are generated using an RGBD camera mounted at the end-effector, pointing towards the table.
The black box simulator $\blackboxsim$ uses Open3D~\cite{open3d} to render RGBD images according to the believed object positions. 

The reward function is defined as
\begin{align*}
    &R_1 = P_{\text{action}} \\
    &R_2 =
    \begin{cases}
        P_{\text{re-observe}} & \text{if } \envMap_s = \envMap_{s'}\\
        0 & \text{otherwise}\\
    \end{cases}\\
    &R_3 =
    \begin{cases}
        R_{\text{terminal}} & \text{if } s' \text{ is terminal}\\
        0 & \text{otherwise}\\
    \end{cases} \\
    &R_{1,2,3} = R_1 + R_2 + R_3 \\
    &R_\text{2D} = R_{1,2,3} + R_{\text{exploration}} + R_{\text{discovery}} \\
    &R_\text{3D} = R_{1,2,3} + R_{\text{refinement}},
\end{align*}

where $P_\text{action}$ is a constant penalty for performing an action, $P_\text{re-observe}$ is a penalty depending on the amount of the environment map that is re-observed when performing an action and $R_\text{terminal}$ is the reward for reaching a terminal state.
The search module uses $R_\text{2D}$ where $R_\text{exploration}$ is a reward for observing cells that are marked as $candidate$, while $R_\text{discovery}$ is a reward for observing objects. 
The pose estimation module uses $R_\text{3D}$, where $R_\text{refinement}$ is a reward for discarding cells marked as $object$ while they are actually $empty$.

\subsection{Evaluations}
\begin{figure*}[ht]
    \centering
    \begin{minipage}[t]{.33\textwidth}
        \centering
        \includegraphics[width=1\textwidth]{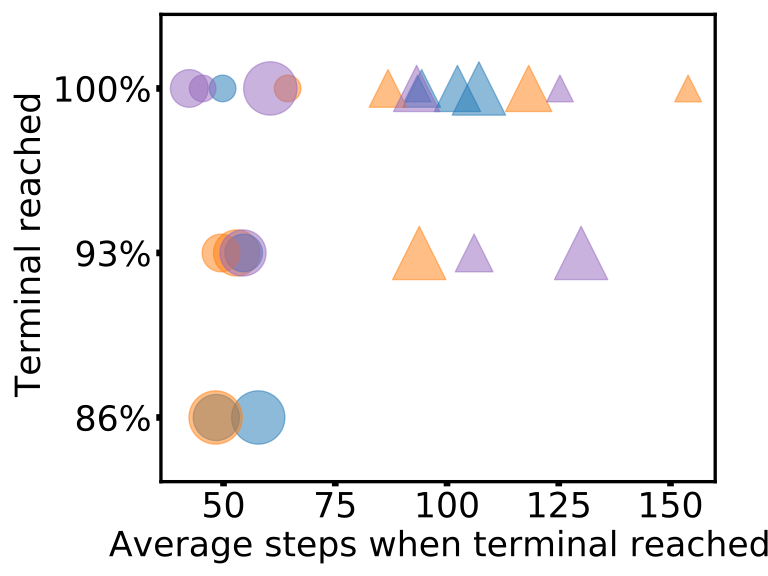}
        \\
        \includegraphics[width=1\textwidth]{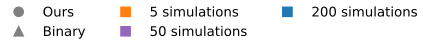} 
        \label{fig:results_obs_type_L}
    \end{minipage}%
    \hfill
    \begin{minipage}[t]{0.33\textwidth}
        \centering
        \includegraphics[width=\textwidth]{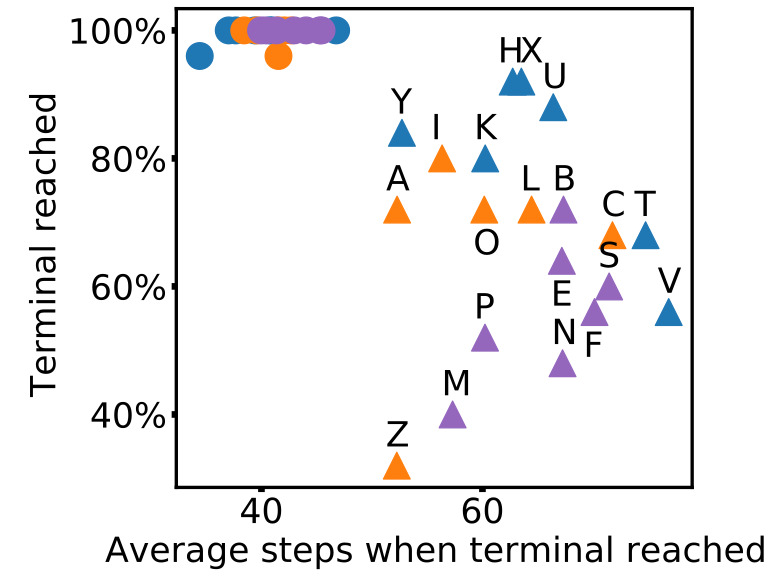}
        \\
        \includegraphics[width=0.6\textwidth]{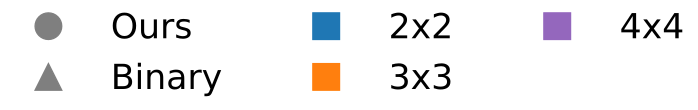}
        \label{fig:results_obs_type_all}
    \end{minipage}%
    \hfill    
    \begin{minipage}[t]{0.33\textwidth}
        \centering
        \includegraphics[width=\textwidth]{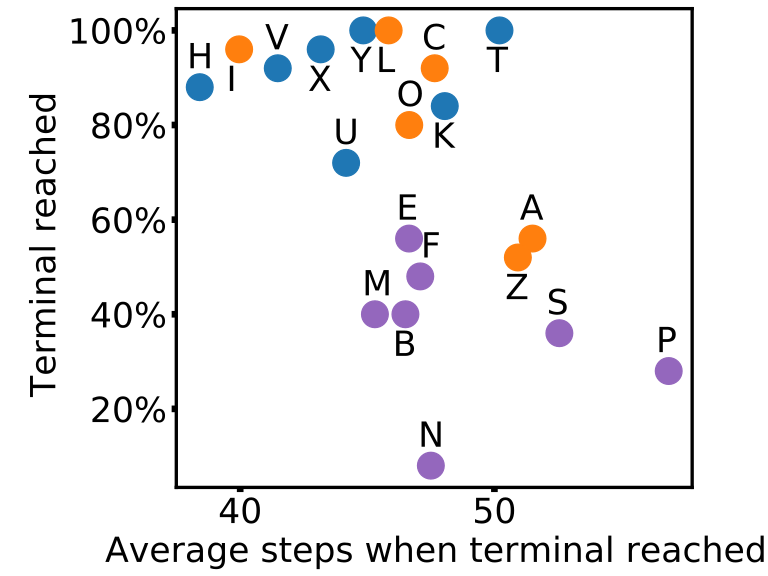}
        \\
        \includegraphics[width=0.6\textwidth]{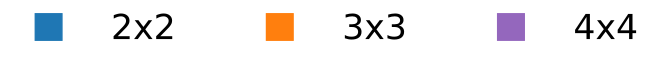}        
        \label{fig:results_avs_ape}
    \end{minipage}    
    
    \caption{\small{
    Comparisons between our mapping from RGB images to observations vs. a binary object detector.
    The number of simulations and particles in the middle and right Figures are fixed to $\nrsimulations=200$ and $\nrparticles=200$, respectively.
    In the legends, the digits near the marker, e.g. $2 \times 2$ indicate the area occupied by the object.
    Left: results for Search Module with object \textit{L}. The marker size is proportional to the number of used particles. The colors indicate the number of simulations.
    Middle: Results for Search Module with all objects and a fixed number of particles and simulations. 
    Right: Results for combining the Search and Pose Estimation Modules with all objects and fixing the number of particles and simulations.
    }}
    \label{different_object_comparison}
    \vspace{-0.5cm}
\end{figure*}

In Fig.~\ref{different_object_comparison} the left and middle figures compare the performance of our search module using binary observations ${\observation \in \{object\_found, \neg object\_found\}}$ against our proposed image to observation mapping, where $\observationVector \in V^{w\times h}$.
The results show that our method is able to find the object faster (less average steps) and more reliably (more times the terminal state is reached).
Therefore, confirming that it is beneficial to introduce a feature extraction map from images to observations, instead of a simple binary classification to decide whether the object is seen or not.

In Fig.~\ref{different_object_comparison} the rightmost plot shows the performance of combining the search and pose estimation modules - AVS + APE.
From the plot is clear that larger objects are harder to handle than smaller ones.
With more maximum number of setups all objects could be potentially estimated, at the expense of higher computation.




\section{Conclusion \& Future Work}
In those situations in which the pose of a certain object is difficult to asses from a single image, actively deciding the actions the agent should do to improve the pose estimation its a mandatory skill for our robots. Nevertheless, solving directly the problem for the objects full pose might be computationally demanding, due to the sensor inputs and the dimension of the possible solutions. To deal with this computational limitation, we propose to split the active pose estimation problem in two stages: A searching process and a pose estimation one. In our work, we propose to model both stage problems as POMDP and solve them with POMCP methods. Framing both stages under the same umbrella allow us to comprenhend how both phases are related to each other and allow us to share information between each other such as the computed belief states. Along the paper, we present two proposed algorithms to integrate POMCP into the search a pose estimation problems and evaluate its performance for a high number of different objects. Through an extensive ablation study, we have evaluated our proposed framework and measure the required number of control steps to solve the task under several variables: different shape and size objects or different type of belief updates.

Our proposed approach allows the use of multiple possible features to represents the belief update metric. Nevertheless, in practise, we limit to the color features. In future works, we aim to explore the performance of the proposed belief update metric for additional features, from handcrafted ones such as edges to learned features. Additionally, we aim to explore the performance of our proposed method in real robot pose estimation tasks and study the required changes to adapt it to a real robot problem.




\bibliographystyle{IEEEtran}
\bibliography{references}

\end{document}